\documentclass[preprint,fleqn,authoryear,10pt]{article}
\usepackage[figuresright]{rotating}
\usepackage{etex}
\usepackage{pstricks-add}
\usepackage[framed,numbered,autolinebreaks,useliterate]{mcode}
\usepackage[hidelinks,dvips]{hyperref}  
\hypersetup{
  colorlinks   = true, 
  urlcolor     = blue, 
  linkcolor    = blue, 
  citecolor   = red 
}
\usepackage[pageref]{backref}
\usepackage{verbatim} 
\usepackage{hypernat}  
\usepackage{mathrsfs}
\usepackage{dsfont} 
\usepackage{amstext}
\usepackage{amsmath}
\usepackage{amsfonts}
\usepackage{cite}
\usepackage{amsthm,amsbsy,bm}
\usepackage{dsfont} 
\usepackage{subeqnarray}
\usepackage{empheq} 
\usepackage{color} 
\usepackage{amssymb}
\usepackage{exscale}  
\usepackage{stmaryrd}  
\usepackage{epsfig}
\usepackage{subfigure}
\usepackage{calc}
\usepackage{amssymb}
\usepackage{amstext}
\usepackage{amsmath}
\usepackage{multicol}
\usepackage{times}%
\usepackage{fancyhdr}
\usepackage[small]{caption}
\usepackage{dsfont} %
\usepackage{subeqnarray}%
\usepackage{empheq} %
\usepackage{color} 
\usepackage{verbatim}
\usepackage{url}
\usepackage{breakurl}%
\usepackage{slashed}  %
\usepackage{exscale}  %
\usepackage{stmaryrd}  %
\usepackage{subeqnarray}
\usepackage{cases}
\newfont{\boldit}{cmbxti10} %
\usepackage{fancyhdr} %
\usepackage{textcomp}
\newtheorem{theorem}{Theorem}

\theoremstyle{definition}
\newtheorem{definition}{Definition}

\theoremstyle{plain}
\theoremstyle{plain}
\swapnumbers \swapnumbers 

\usepackage{amsmath,varwidth,array,ragged2e,tabularx}
\usepackage{tikz}

\usepackage{graphicx} %
\usepackage{multirow,makecell,rotating} 
\newcolumntype{L}{>{\varwidth[c]{\linewidth}}l<{\endvarwidth}}
\newcolumntype{M}{>{$}l<{$}}
\newcommand{\tc}[1]{\multicolumn{1}{c}{#1}} 
\renewcommand*{\backref}[1]{} 
\renewcommand*{\backrefalt}[4]{%
    \ifcase #1 (Not cited.)%
    \or        (Cited on page~#2.)%
    \else      (Cited on pages~#2.)%
    \fi}
\allowdisplaybreaks[3]

\makeatletter
\def\hlinewd#1{%
\noalign{\ifnum0=`}\fi\hrule \@height #1 %
\futurelet\reserved@a\@xhline}

\makeatother

\makeatletter
\def\@cite#1#2{\textsuperscript{[{#1\if@tempswa , #2\fi}]}}
\makeatother

\makeatletter
  \def\my@tag@font{\normalsize}
  \def\maketag@@@#1{\hbox{\m@th\normalfont\my@tag@font#1}}
  \let\amsmath@eqref\eqref
  \renewcommand\eqref[1]{{\let\my@tag@font\relax\amsmath@eqref{#1}}}
\makeatother

\begin{document}
\title{A General Homogeneous Matrix Formulation to 3D Rotation Geometric Transformations}
\author{F. LU$^1$,  Z. CHEN$^2$\thanks{Corresponding author}\\
}
\date{}
\maketitle

\begin{abstract}
{\normalsize\noindent  We present algebraic projective geometry definitions of 3D rotations so as to bridge a small gap between the applications and the definitions of 3D rotations in homogeneous matrix form. A general homogeneous matrix formulation to 3D rotation geometric transformations is proposed which suits for the cases when the rotation axis is unnecessarily through the coordinate system origin given their rotation axes and rotation angles.

General three-dimensional rotation formula~\eqref{eqn:3D homogeneous roation} and~\eqref{eqn:3D rotation matrix vector Euclidean} similar to the Euler-Rodrigues formula were presented. The matrix-vector form of 3D rotation in Euclidean space is especially suited for numerical applications where gimbal lock is a concern.}
\end{abstract}


\section{Introduction}

Geometric transformation rotation is a basic and fundamental concept which has applications in computer graphics, vision and robotics and has been investigated and depicted thoroughly in many classic literatures~\cite{CAD,Goldman,Perspectives,Salomon,VinceFormulae,VinceMath,VinceRotation}. Rotations of practical importance are those 2D and 3D rotation transformations represented by quaternion and vectors in Euclidean space, and by homogeneous matrices in projective spaces.

It is well known {\em quaternion} is a useful tool in representing 3D rotations~\cite{Salomon,VinceFormulae,VinceMath,VinceRotation} which, however, has the difficulty of representing general 3D rotations with rotation axes not passing through the coordinate system origin. An alternative Rodrigues formula which explicitly contains the point vector to be transformed can be used to solve this problem~\cite[p.165]{Goldman}. 

Since all 3D rotations thus defined are actually dependent on their Euclidean geometric meaning, therefore their homogeneous forms are actually lack of rigorous definition from the viewpoint of incidence or projective geometry. We admit that the final formulation presented by us in this article can also be obtained by the conventional approaches, e.g., by rewriting Euler-Rodrigues rotation~\cite[pp583-585]{Hartley} even more conveniently without cumbersome manual symbolic computation and symbolic simplification. While traditional methods excel at producing analytic expressions, they leave a significant gap in the understanding of the theoretical underpinnings of the geometric transformations framework~\cite{investigation,meaning,lu2013unified}.  Therefore, we contend that a rigorous, algebraic approach, even if these involve tedious symbolic calculations for the derivation of definitions and representations, is essential for constructing a robust theoretical framework.

To the best of our knowledge, a rigorous, general definition of general 3D rotations in homogeneous matrix form yet from the viewpoint of incidence geometry that satisfies the aforementioned criteria is currently unavailable in the literature. Despite the potential for controversy surrounding this argument, we suggest deferring these discussions and continuing our investigation. This approach parallels our previous work, where we reformulated central projection, parallel projection, and other geometric transformations and projections as Householder elementary matrices~\cite{investigation,meaning,lu2013unified}. In order to represent general 3D rotations in homogeneous matrices, 3D rotations first have to be well-defined. 

Due to the inherent Euclidean geometric intuition regarding geometric transformations, it is easy to underestimate the necessity of rigorously defining homogeneous geometric transformations algebraically in projective or incidence geometry. The following examples demonstrate the necessity of rigorously redefining rotation.

Given the two homogeneous matrices in Equation~\eqref{eqn:3D rotation example}, the new framework proposed~\cite{investigation,meaning,lu2013unified} without a definition to rotations may face logical challenges in distinguishing between a rotation (\eqref{eqn:3D rotation example a}) and a non-rotation (\eqref{eqn:3D rotation example b}) without relying on traditional geometric concepts like distance and perpendicularity.
\begin{subequations}
\label{eqn:3D rotation example}
\begin{align}\label{eqn:3D rotation example a}
\left(
\begin{array}{cccc}
 50+10 \sqrt{3} & -30+15 \sqrt{3}-\sqrt{35} & -10+5 \sqrt{3}+3 \sqrt{35} & 40-20 \sqrt{3}+4 \sqrt{35} \\
 -30+15 \sqrt{3}+\sqrt{35} & 18+26 \sqrt{3} & 6-3 \sqrt{3}+5 \sqrt{35} & 88-44 \sqrt{3}+4 \sqrt{35} \\
 -10+5 \sqrt{3}-3 \sqrt{35} & 6-3 \sqrt{3}-5 \sqrt{35} & 2+34 \sqrt{3} & -64+32 \sqrt{3}+8 \sqrt{35} \\
 0 & 0 & 0 & 70 \\
\end{array}
\right)
\end{align}
\begin{align}\label{eqn:3D rotation example b}
\left(
\begin{array}{cccc}
 -11 & -32 & 11 & 60 \\
 8 & 20 & -2 & -24 \\
 0 & 0 & 6 & 0 \\
 0 & 0 & 0 & 6 \\
\end{array}
\right) \end{align}
\end{subequations}

The purely algebraic nature of homogeneous matrices in projective or incidence geometry necessitates a definition of geometric transformations that is both intuitive and independent of Euclidean concepts. In this context, Euclidean notions such as {\em distance} are no longer applicable, {\em angles} are defined using Laguerre's formula~\cite[pp.342,409]{Perspectives}, and Euclidean transformations become undefined. A {\em well-defined} geometric transformation should allow for a bidirectional mapping: from a homogeneous matrix to its geometric interpretation, and conversely, from geometric properties to a homogeneous matrix representation.


Additionally, in order to obtain reference frame independent matrix of a 3D rotation, the conventional representation highly depends on the following unarticulated truth, which holds for any geometric transformation $\mathscr{T}_0$ in square matrices({Note: column vector convention used unless otherwise specified}):
\begin{theorem}\label{zeroth}
Suppose $\mathscr{T}_0$ is a geometric transformation in projective space which transforms an arbitrary point $X$ into $Y$; and the homogeneous coordinates of $X$ and $Y$ in reference coordinate systems $(I)$ and $(II)$ are $(x)$,\;$(y)$,\;$(x')$,\;$(y')$\; respectively; the transformation matrices of $\mathscr{T}_0$ in $(I)$ and $(II)$ are ${\boldsymbol A}$ and ${\boldsymbol B}$ respectively, i.e., $(y)={\boldsymbol A}(x)$, $(y')={\boldsymbol B}(x')$; suppose the coordinate transformation from $(I)$ to $(II)$ is a nonsingular square matrix ${\boldsymbol T}$, i.e., $(x')={\boldsymbol T}(x)$, $(y')={\boldsymbol T}(y)$;  then:
\begin{align*}
(y')\,=\,{\boldsymbol B}(x')\,=\,{\boldsymbol T}\,(y)\,=\,{\boldsymbol T}\,{\boldsymbol A} \,(x)\,=\,{\boldsymbol T} \,{\boldsymbol A} \,{\boldsymbol T}^{-1} \,(x') \quad \forall X, Y \Rightarrow {\boldsymbol B}\,=\,{\boldsymbol T}\,{\boldsymbol A}\,{\boldsymbol T}^{-1}
\end{align*}
 The matrices of $\mathscr{T}_0$ in  $(I)$ and $(II)$  are similar.
\end{theorem}

Though theorem~\ref{zeroth} indicates that the characteristic algebraic features of a homogeneous geometric transformation are its eigenvalues and their algebraic and geometric multiplicities, none of the conventional definitions of geometric transformations has taken advantage of such rules to reveal the inherent connection between the geometric meaning of homogeneous matrices and their eigenvalues.

Rotation matrices are similar to Givens rotation matrices, differing only by orthogonal factors. This suggests that such matrices form the bedrock of traditional rotation definitions, including those used in the Euler-Rodrigues formulation.

Take rotation with $\theta$ angle and axis passing through $P:(2,1,5,1)^T$ and $Q:(4,7,2,1)^T$ in homogenous coordinates as an example~\cite[p.47]{CAD}. {\color{black}Note that we use {\em row vector} convention only in this example such that the results here without transposing are consistent with those in}~\cite[p.47]{CAD}. The method tries to construct a series of {\em Euclidean geometric transformations} $E_i(i=1,2,\cdots,n)$ such that the final rotation $R$  obtained in~\eqref{eqn:rotation example} per equation~\eqref{eqn:Givens rotation} is similar to a known rotation $R_0$ in~\eqref{eqn:Givens rotation} in the standard Givens rotation form  around coordinate axis $z$:
\begin{equation}\label{eqn:Givens rotation}
R_0=\left(
\begin{array}{cccc}
 \cos \theta & \sin\theta & 0 & 0 \\
 -\sin\theta & \cos\theta & 0 & 0 \\
 0 & 0 & 1 & 0 \\
 0 & 0 & 0 & 1 \\
\end{array}
\right); \quad R = {\color{blue}E_n^{-1} \cdots E_2^{-1}\cdot E_1^{-1} \cdot} R_0 {\color{blue}\cdot E_1\cdots E_2\cdot E_n}
\end{equation}

\begin{equation}\label{eqn:rotation example}
R=\dfrac{1}{49}\left(
\begin{array}{cccc}
45\cos\theta+4&12-12\cos\theta-21\sin\theta&6\cos\theta-42\sin\theta-6&0\\[5pt]
12-12\cos\theta+21\sin\theta&36+13\cos\theta&18\cos\theta+14\sin\theta-18&0\\[5pt]
6\cos\theta+42\sin\theta-6&18\cos\theta-14\sin\theta-18&40\cos\theta+9&0\\[5pt]
108-108\cos\theta-231\sin\theta&79-79\cos\theta+112\sin\theta&230-230\cos\theta+70\sin\theta&49\\
\end{array}
\right)
\end{equation}

Such an approach in determining the homogeneous matrix of a general rotation has the following drawbacks: (i)~The definition of a general rotation in homogeneous representation and projective space is algebraically dependent on the standard Givens rotations inherited from Euclidean spaces; (ii)~A series of Euclidean transformations $E_i$, the homogeneous matrices of which are algebraically undefined in projective geometry, have to be employed and sometimes chosen arbitrarily, which makes the procedure complicate to program and code; (iii)~There is no algebraic definition in projective space to determine the geometric meaning of homogeneous matrix in equation~\eqref{eqn:rotation example} conversely without using such non-projective-geometry concepts as distance.

By using an extended Desargues theorem~\cite[pp.75~\texttildelow~76]{Veblen} and examining the thus obtained extended Desarguesian configuration~\cite{investigation,meaning,lu2013unified} via an algebraic projective geometric approach, Householder's elementary matrices~\cite[pp.1~\texttildelow~3]{Householder} were rewritten into new forms as in table~\ref{classification table} and defined as {\em stereohomology} which consists of most of the basic geometric transformations with such {\em nice} definitions in projective space. 

In this work, we will extend such work to rotations which are not any more {\em elementary} to solve the definition issue for 3D rotations in projective space. General homogeneous 3D rotations will also be presented. We first define rotations in 2D and 3D projective spaces, and then present two approaches to obtain homogeneous matrix formula of general 3D rotations, i.e., the rotation axes of them do not have to pass through the {\em coordinate system origin}.

\section{Definition of 2D and 3D rotations}

Note that {\em column} homogeneous vectors, instead of {\em row} vectors,  will be employed as default point and geometric transformation representation convention hereafter in this paper, which is different from the convention adopted by~\cite[p.47]{CAD}.

The uncertainty of representing an arbitrary 3D rotation in homogeneous matrix form mainly lies in the 3D rotation axis representation. Though lines in 3-space can be represented in their Pl\"{u}ker coordinates \cite[pp.68~\texttildelow~72]{Hartley} \cite[pp.216~\texttildelow~218]{Perspectives}, which brings convenience in representing intersection or joining of lines with 3D points and hyperplanes, it is difficult to use the coordinates information directly into geometric transformation representation.

Though quaternions can elegantly express rotations with arbitrary axis passing through the origin in three dimensions \cite[pp.50~\texttildelow~54]{Salomon} \cite[pp.177~\texttildelow~180]{VinceRotation},\cite[pp.92~\texttildelow~95]{VinceMath}, for rotation axes which are in general not passing through the origin, we may have to use a pair of translations which are inverse to each other so that to obtain the desired rotation per theorem \ref{zeroth}.

A rotation can be defined as the compound operation or the product of two reflections according to~\cite[pp.419~\texttildelow~422]{Perspectives}, but there is no simple rotation representation derived based on such definitions yet. In this section we shall both use the compound transformation of two orthographic {\em reflections} defined as  involutory stereohomology in table~\ref{classification table} and use the eigen-system of the rotation which is inherent algebraic features per theorem~\ref{zeroth}, to represent a general rotation.

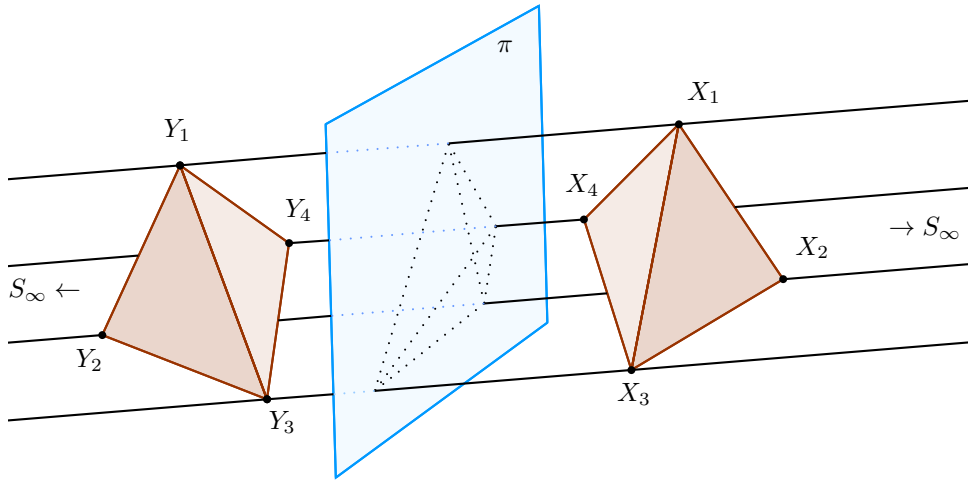
\begin{figure}[!hpt]
\begin{center}
\newrgbcolor{zzttqq}{0.6 0.5 0.3}
\newrgbcolor{zzttql}{0.2 0.1 0}
\newrgbcolor{qqzzff}{0 0.6 1}
\newrgbcolor{wwzzff}{0.4 0.6 1}
\newrgbcolor{zzccff}{0.6 0.8 1}
\psset{xunit=0.75cm,yunit=0.75cm,algebraic=true,dotstyle=o,dotsize=3pt 0,linewidth=0.8pt,arrowsize=3pt 2,arrowinset=0.25}
\begin{pspicture*}(-3,-4)(14,6)
\pspolygon[linecolor=zzttqq,fillcolor=zzttqq,fillstyle=solid,opacity=0.2](8.88,3.19)(10.73,0.44)(8.04,-1.17)
\pspolygon[linecolor=zzttqq,fillcolor=zzttqq,fillstyle=solid,opacity=0.1](8.88,3.19)(7.2,1.5)(8.04,-1.17)
\pspolygon[linecolor=zzttqq,fillcolor=zzttqq,fillstyle=solid,opacity=0.2](0.04,2.46)(-1.34,-0.55)(1.58,-1.69)
\pspolygon[linecolor=zzttqq,fillcolor=zzttqq,fillstyle=solid,opacity=0.1](0.04,2.46)(1.97,1.08)(1.58,-1.69)
\pspolygon[linecolor=qqzzff,fillcolor=qqzzff,fillstyle=solid,opacity=0.05](2.62,3.19)(6.4,5.29)(6.55,-0.33)(2.8,-3.08)
\psline[linecolor=zzttql](8.88,3.19)(10.73,0.44)
\psline[linecolor=zzttql](10.73,0.44)(8.04,-1.17)
\psline[linecolor=zzttql](8.04,-1.17)(8.88,3.19)
\psline[linecolor=zzttql](8.88,3.19)(7.2,1.5)
\psline[linecolor=zzttql](7.2,1.5)(8.04,-1.17)
\psline[linecolor=zzttql](8.04,-1.17)(8.88,3.19)
\psline[linecolor=zzttql](0.04,2.46)(-1.34,-0.55)
\psline[linecolor=zzttql](-1.34,-0.55)(1.58,-1.69)
\psline[linecolor=zzttql](1.58,-1.69)(0.04,2.46)
\psline[linecolor=zzttql](0.04,2.46)(1.97,1.08)
\psline[linecolor=zzttql](1.97,1.08)(1.58,-1.69)
\psline[linecolor=zzttql](1.58,-1.69)(0.04,2.46)
\psline[linecolor=qqzzff](2.62,3.19)(6.4,5.29)
\psline[linecolor=qqzzff](6.4,5.29)(6.55,-0.33)
\psline[linecolor=qqzzff](6.55,-0.33)(2.8,-3.08)
\psline[linecolor=qqzzff](2.8,-3.08)(2.62,3.19)
\psline[linestyle=dotted](4.8,2.85)(5.64,1.38)
\psline[linestyle=dotted](5.64,1.38)(5.43,0.01)
\psline[linestyle=dotted](3.49,-1.54)(5.43,0.01)
\psline[linestyle=dotted](4.8,2.85)(3.49,-1.54)
\psline[linestyle=dotted](4.8,2.85)(5.43,0.01)
\psline[linestyle=dotted](5.64,1.38)(3.49,-1.54)
\psline(5.64,1.38)(7.2,1.5)
\psline(4.8,2.85)(8.88,3.19)
\psline(3.49,-1.54)(8.04,-1.17)
\psline(5.43,0.01)(7.61,0.19)
\psline(1.58,-1.69)(2.76,-1.6)
\psline(1.78,-0.29)(2.72,-0.21)
\psline(1.97,1.08)(2.68,1.13)
\psline(0.04,2.46)(2.63,2.68)
\psline(-6.01,-2.31)(1.58,-1.69)
\psline(-6.16,-0.94)(-1.34,-0.55)
\psline(-6.28,1.95)(0.04,2.46)
\psline(-6.36,0.4)(-0.7,0.86)
\psline(8.88,3.19)(18.18,3.94)
\psline(8.04,-1.17)(18.67,-0.3)
\psline(10.73,0.44)(18.6,1.08)
\psline(9.87,1.72)(18.42,2.42)
\psline[linestyle=dotted,linecolor=wwzzff](2.63,2.68)(4.8,2.85)
\psline[linestyle=dotted,linecolor=wwzzff](2.68,1.13)(5.64,1.38)
\psline[linestyle=dotted,linecolor=wwzzff](2.72,-0.21)(5.43,0.01)
\psline[linestyle=dotted,linecolor=zzccff](2.76,-1.6)(3.49,-1.54)
\rput[tl](5.66,4.67){$\pi$}
\rput[tl](12.6,1.55){$\to S_{\infty}$}
\rput[tl](-3,0.45){$ S_{\infty} \leftarrow$}
\psdots[dotstyle=*](8.88,3.19)
\rput[bl](9.03,3.56){$X_1$}
\psdots[dotstyle=*](10.73,0.44)
\rput[bl](10.95,0.81){$X_2$}
\psdots[dotstyle=*](8.04,-1.17)
\rput[bl](7.79,-1.79){$X_3$}
\psdots[dotstyle=*](7.2,1.5)
\rput[bl](6.87,1.92){$X_4$}
\psdots[dotstyle=*](0.04,2.46)
\rput[bl](-0.25,2.89){$Y_1$}
\psdots[dotstyle=*](-1.34,-0.55)
\rput[bl](-1.81,-1.19){$Y_2$}
\psdots[dotstyle=*](1.58,-1.69)
\rput[bl](1.61,-2.28){$Y_3$}
\psdots[dotstyle=*](0.04,2.46)
\psdots[dotstyle=*](1.97,1.08)
\rput[bl](1.9,1.48){$Y_4$}
\psdots[dotstyle=*](1.58,-1.69)
\end{pspicture*}
\end{center}
\caption{Extended Desarguesian configuration for reflection}
\label{fig:Desarguesian configuration reflection}
\end{figure}

The definition of an orthographic reflection in~\cite{meaning,lu2013unified} takes advantage of the existence and uniqueness of an involutory projective transformation which transforms $X_i$ and $S$ in the extended Desargues configuration $X_1X_2X_3X_4-S-Y_1Y_2Y_3Y_4$ (as in figure~\ref{fig:Desarguesian configuration reflection}) into $Y_i$ and $S$ in sequence respectively. The homogeneous square matrix formulation of such a reflection was proved to be in the form as indicated in table~\ref{classification table}~\cite{lu2013unified}.

Note that in order to make the definitions in algebraic projective geometry {\em compatible with} the Euclidean geometry intuitions in one's mind, we have to make choices to distinguish regular and infinite geometric elements which are algebraically indistinguishable in projective space. Then we redefine homogeneous rotations in ${\mathbb P}^n$ ( only when $n$ = 2,3) in this paper as:

\begin{definition}[Rotation]\label{rotation1}
A rotation in ${\mathbb P}^n$ is a compound transformation of two orthographic reflections of which the stereohomology centers $S_1$ and $S_2$ are different infinite points, and sterehomology hyperplanes $\pi_1$ and $\pi_2$ are regular elements(see table~\ref{classification table} for definitions of elementary geometric transformations).

The rotation angle $\theta$ of the rotation is twice that of dihedral angle $\omega$ between $\pi_1$ and $\pi_2$ which can be represented by Laguerre's formula by involving cross ratio~\cite[pp.342,409]{Perspectives}.

\end{definition}

The above definition~\ref{rotation1} is directly borrowed from the classic definitions in projective geometry, and is theoretically dependent on the possibility of defining {\em normal reflection} as an involutory stereohomology in table~\ref{classification table}, detailed illustation of which can be seen in~\cite{investigation,meaning}(in Chinese) and~\cite{lu2013unified} where modified Householder's elementary matrices~\cite[1\texttildelow 3]{Householder} are presented and defined into {\em stereohomology} as in table~\ref{classification table} based on an extension to Desargues theorem~\cite[75\texttildelow 76]{Veblen}. Otherwise, we have not find any other opportunity of define rotation via such an approach in algebraic projective geometry. It is only based on definition~\ref{rotation1} that we can obtain a pure algebraic definition~\ref{rotation2} of 2D and 3D rotations in projective space without using any non-projective-geometry concept, i.e., it is logically inappropriate to immediately adopt a Givens matrix as a {\em standard} rotation. 

\begin{definition}[Rotation]\label{rotation2}
A rotation in ${\mathbb P}^n$($n$ = 2,3) with  rotation angle $\theta$  and rotation axis $l$ (the latter of which should be able to be represented as the intersection of two hyperplanes in ${\mathbb P}^n$) is a projective transformation of which:

(1) the ratios of all eigenvalues are cos$\theta \pm i\cdot$ sin$\theta $ and 1;

(2) points on rotation axis $l$ are the associated eigenvectors with the ratio 1 real eigenvalue, and

(3) the associated eigenvectors with eigenvalues of ratios cos$\theta \pm i\cdot$ sin$\theta $ are the intersetion points of the imaginary conics\cite[p.204]{Vaisman} and the infinite hyperplane in ${\mathbb P}^2$, or are the intersection points of the imaginary quadrics\cite[p.204]{Vaisman}, the infinite hyperplane, and any regular hyperplane of which the normal direction is the direction of the rotation axis when it is in ${\mathbb P}^3$;

(4) it is a distance preserving transformation, i.e., the transpose of the upper-left $3\times 3$ submatrix is also the submatrix' inverse up to a non-zero factor.
%
%
%

\end{definition}

We shall obtain homogeneous rotations based on definitions \ref{rotation1} and \ref{rotation2} via two approaches different from those in~\cite[pp.33~\texttildelow~34,43~\texttildelow~48]{CAD}, \cite[pp.43~\texttildelow~52]{Salomon}, \cite[p.36]{VinceFormulae}, \cite[pp.89~\texttildelow~90,115~\texttildelow~118,177~\texttildelow~180]{VinceRotation}:
\begin{description}
\item[(I)] find two hyperplanes of which their intersection line being rotation axis and the dihedral angle $\omega$ being half the angle $\theta$, then the products of reflections about the two hyperplanes will be desired rotation and its inverse (per definition \ref{rotation1}), further characteristic geometric features of positive direction of rotation axes and the right- or left-handed rule, can finalize the desired rotation;
\item[(II)] find all the eigenvalues and their associate eigenvectors, then the rotation and its inverse can be obtained by reconstructing from its eigen-decomposition factors(per definition \ref{rotation2}), further characteristic geometric information on the rotation similar to above uniquely determines the rotation.
\end{description}

\section{Formulation to general 3D rotations}
The formulation problem of 2D rotations has actually been well resolved~\cite{CAD,Salomon,VinceFormulae,VinceMath,VinceRotation} even without considering the new definitions~\ref{rotation1} and~\ref{rotation2} here. Now let's go on with the 3D homogeneous rotations. The major difference from 2D cases is that the 3D rotation axis can be represented as either intersection of two hyperplanes or a line joining two points. For the latter, there are two possibilities: (1) both points are regular; (2) one point is infinite and represents direction of the axis.

When rotation axis is determined by two hyperplanes, we use the first approach by definition \ref{rotation1}. The key is to find a pair of hyperplanes, the intersection line of which is the rotation axis, and the dihedral angle of which is half the rotation angle $\theta$.

If the rotation axis is given by the intersection of two hyperplanes $\pi_1$ and $\pi_2$\cite[pp.58~\texttildelow~69]{Vaisman}:
\begin{equation}\label{3D axis by hyperplanes}
\left\{ {\begin{array}{*{20}{c}}
{{\pi _1}:{a_1} \cdot x + {b_1} \cdot y + {c_1} \cdot z + d_1 = 0},\quad\text{where } {a_1^2 + b_1^2 + c_1^2}=1 \\
{{\pi _2}:{a_2} \cdot x + {b_2} \cdot y + {c_2} \cdot z + d_2 = 0}, \quad\text{where } {a_2^2 + b_2^2 + c_2^2}=1
\end{array}} \right.
\end{equation}
Suppose the dihedral angle between  $\pi_1$ and $\pi_2$ is $\omega$, then hyperplane equations which intersect $\pi_1$ and $\pi_2$ at the rotation axis and have $\pm \theta \slash 2$ dihedral angles with $\pi_1$ are:

\begin{equation}\label{3D rotation hyperplanes}
\left\{ {\begin{array}{*{20}{c}}
{{\pi _{\omega  - \frac{\theta }{2}}}:\sin \left( {\omega  - \dfrac{\theta }{2}} \right) \cdot \dfrac{{\left( {{a_1} \cdot x + {b_1} \cdot y + {c_1} \cdot z + {d_1}} \right)}}{{\sqrt {a_1^2 + b_1^2 + c_1^2} }} + \sin \dfrac{\theta }{2} \cdot \dfrac{{\left( {{a_2} \cdot x + {b_2} \cdot y + {c_2} \cdot z + {d_2}} \right)}}{{\sqrt {a_2^2 + b_2^2 + c_2^2} }} = 0}\\[25pt]
{{\pi _{\omega  + \frac{\theta }{2}}}:\sin \left( {\omega  + \dfrac{\theta }{2}} \right) \cdot \dfrac{{\left( {{a_1} \cdot  x + {b_1} \cdot y + {c_1} \cdot z + {d_1}} \right)}}{{\sqrt {a_1^2 + b_1^2 + c_1^2} }} - \sin \dfrac{\theta }{2} \cdot \dfrac{{\left( {{a_2} \cdot x + {b_2} \cdot y + {c_2} \cdot z + {d_2}} \right)}}{{\sqrt {a_2^2 + b_2^2 + c_2^2} }} = 0}
\end{array}} \right.
\end{equation}

By using either of the two hyperplane equations in equation \eqref{3D rotation hyperplanes} and that of $\pi_1$, two reflections can easily be obtained, the product of which are the desired 3D homogeneous rotation and its inverse. Generally equation \eqref{3D rotation hyperplanes} is more suitable for numerical rotation matrix estimation since the two hyperplanes can be arbitrary and the rotation matrix thus obtained is not unique and therefore the analytical form of the rotation matrix will not be given here.

Next let us consider the rotation axis determined by two points. Since when both the two points are regular, the direction (per definition of {\em direction} as {\em stereohomology} in table~\ref{classification table}) of one point from another will be the direction of the axis, we only discuss the case when one is an regular point while another is infinite, i.e., a direction. Denote the regular point as ($x_0,y_0,z_0,1$)$^T$ and the direction ($a,b,c,0$)$^T$(without loss of generality, let $a^2+b^2+c^2=1$).

If the representation of axis by two points can be converted into by two hyperplanes similar to that in equation \eqref{3D axis by hyperplanes} then we can use the similar approach by equations \eqref{3D axis by hyperplanes} and \eqref{3D rotation hyperplanes} to obtain the rotation. The pencil of hyperplanes through the line joining  ($x_0,y_0,z_0,1$)$^T$ and ($a,b,c,0$)$^T$ should satisfy:
\begin{equation}
\begin{aligned}
\left\{ \begin{aligned}\label{plane pencil through 2 points}
\alpha \cdot\left(x - x_0\right)\hphantom{A}& + & \beta  \cdot \left(y - y_0\right)\hphantom{A}& + & \gamma  \cdot \left(z -z_0\right) \hphantom{A}&=& 0\\
\alpha \cdot a \hphantom{AAA}& + & \beta  \cdot b \hphantom{AAA}& + & \gamma  \cdot c \hphantom{AAA}& = & 0
\end{aligned}\right.  \hphantom{AAAAAAAAAAAAAAAA}\\
 {\textrm{where}} \left(\alpha, \beta, \gamma,0 \right)^T {\textrm{ is the normal direction of the hyperplane pencil passing through the two points}}.
\end{aligned}
\end{equation}

Without loss of generality, let $c\neq 0$, substitute $\gamma  = -(\alpha  \cdot a + \beta  \cdot b)\slash c$ into equation \eqref{plane pencil through 2 points} we have:

\[\alpha  \cdot c \cdot \left( {x - {x_0}} \right) + \beta  \cdot c \cdot \left( {y - {y_0}} \right) - \left( {\alpha  \cdot a + \beta  \cdot b} \right) \cdot \left( {z - {z_0}} \right) = 0 ,\]
which can be re-written into:
\begin{equation}\label{plane intersection of 2 points}
\alpha \cdot \left( {c \cdot x + 0 \cdot y - a \cdot z + \left( {a{z_0} - c{x_0}} \right)} \right) + \beta\cdot \left( {0 \cdot x + c \cdot y - b \cdot z + \left( {b \cdot {z_0} - c \cdot {y_0}} \right)} \right) = 0
\end{equation}

By obtaining equation \eqref{plane intersection of 2 points}, we have successfully represented the rotation axis determined by two points into the a pencil of hyperplanes passing through the axis\cite[pp.58~\texttildelow~69]{Vaisman}. Then use the first approach as used by equations \eqref{3D axis by hyperplanes} and \eqref{3D rotation hyperplanes} we can obtain the 3D homogeneous rotation.

Now let's try the second approach when rotation axis is determined by ($x_0,y_0,z_0,1$)$^T$ and ($a,b,c,0$)$^T$. Clearly both ($x_0,y_0,z_0,1$)$^T$ and ($a,b,c,0$)$^T$ are eigenvectors associated with eigenvalue 1. As eigenvectors, imaginary and infinite points ($x_1,x_2,x_3,x_4$)$^T$ satisfy:
 \begin{equation}\label{3D eigenvector constraints}\left\{ {\begin{aligned}
{ x_1^2+x_2^2+x_3^2+x_4^2 = 0}\\
{ x_4 = 0}\\
{a\cdot x_1+b\cdot x_2+c\cdot x_3 + d\cdot x_4 = 0 }
\end{aligned}} \right. ,\quad \textrm{where $d$ is arbitrary since $x_4=0$.} \end{equation}

Without loss of generality, let the two conjugated eigenvectors are:

$\hphantom{AAAAAAAAAAA}$ ($\alpha + i\cdot \beta$, $\lambda + i\cdot \rho$, 1, 0)$^T$ $\hphantom{AA}$ and $\hphantom{AA}$  ($\alpha - i\cdot \beta$, $\lambda - i\cdot \rho$, 1, 0)$^T$,

\noindent which satisfy equations in \eqref{3D eigenvector constraints}, and therefore are rewritten as in equations \eqref{complex eigenvector equations} on the left and can be solved as those in equations \eqref{complex eigenvector equations} on the right:

\begin{flalign}\label{complex eigenvector equations}
{\text{Equations:} }
\left\{ {\begin{array}{*{20}{c}}
{{\alpha ^2} - {\beta ^2} + {\lambda ^2} - {\rho ^2} + 1 = 0}\\
{\alpha \beta  + \lambda \rho  = 0}\\
{a\alpha  + b\lambda  + c = 0}\\
{a\beta  + b\rho  = 0}
\end{array}} \right.
&{\textrm {Solution:\quad}}
\left\{ {\begin{array}{*{20}{c}}
{\alpha  = \dfrac{{ - ac}}{{{a^2} + {b^2}}}}\\[10pt]
{\beta  = \dfrac{{ \pm b\sqrt {{a^2} + {b^2} + {c^2}} }}{{{a^2} + {b^2}}}}\\[10pt]
{\lambda  = \dfrac{{ - bc}}{{{a^2} + {b^2}}}}\\[10pt]
{\rho  = \dfrac{{ \mp a\sqrt {{a^2} + {b^2} + {c^2}} }}{{{a^2} + {b^2}}}}
\end{array}} \right.
\end{flalign}

The right-handed 3D homogeneous rotation by the four eigenvalues and their associated eigenvectors is given as in~\eqref{3D homogeneous roation}, which for application convenience has been rewritten into a user friendly form similar to the classic Rodrigues' formula~\cite[p.165]{Goldman} 
({\color{black}Note}: without loss of generality, we assumed $\color{red}a^2+b^2+c^2=1$):
{
\begin{gather}\label{3D homogeneous roation}
\boldsymbol{R}^{3D}\left(x_0,y_0,z_0,a,b,c,\theta\right)=\mathscr{C}_1+\left(\sin\theta\cdot\mathscr{A}_2- \left(1-\cos\theta\right)\cdot\mathscr{O}_3\right)\cdot \mathscr{T}_4
\end{gather}
}
where:
\begin{flalign*}
\quad\mathscr{C}_1=
\begin{array}{c}
\underbrace{\left[
\begin{array}{*{20}{c}}
1&0&0&0\\[2pt]
0&1&0&0\\[2pt]
0&0&1&0\\[2pt]
0&0&0&{2 - \cos \theta }
\end{array} \right]} \\ \text{Central dilation}{_{\tiny\tfrac{1}{ (2-\cos\theta)}}}
\end{array},  & \qquad\quad\mathscr{A}_2  =  {\begin{array}{c}
\underbrace{
\begin{array}{c}
 {{\left[ {\begin{array}{*{20}{c}}
{\color{blue}0}&{\color{blue}-c}&{\color{blue}b}&0\\[2pt]
{\color{blue}c}&{\color{blue}0}&{\color{blue}-a}&0\\[2pt]
{\color{blue}-b}&{\color{blue}a}&{\color{blue}0}&0\\[2pt]
0&0&0&0
\end{array}} \right]}}
\end{array}
} \\ \text{Antisymmetric matrix}
\end{array}}
\end{flalign*}

\begin{flalign*}
\quad\mathscr{O}_3= \begin{array}{c}
\underbrace{  {I - {\left[ {\begin{array}{*{20}{c}}
a\\[2pt]
b\\[2pt]
c\\[2pt]
0
\end{array}} \right] \cdot \left[ {\begin{array}{*{20}{c}}
a&b&c&0
\end{array}} \right]}} }\\
\text{Orthographic parallel projection} \end{array}, \quad & \mathscr{T}_4=\begin{array}{c}
\underbrace{\left[ {\begin{array}{*{20}{c}}
1&0&0&{ - {x_0}}\\[2pt]
0&1&0&{ - {y_0}}\\[2pt]
0&0&1&{ - {z_0}}\\[2pt]
0&0&0&1
\end{array}}  \right]}\\ \text{Translation}\end{array}
\end{flalign*}

And now it is easy to identify the homogeneous matrix in~\eqref{eqn:3D rotation example a} per definition~\ref{rotation2} as a rotation by eigen-decomposition: the rotation axis has a direction of $(-5, 3, 1, 0)^T$(choice of positive direction) while passing through fixed points as $({-4-5t, 4+3t, t, 1})^T\;\left(\forall t\in\mathbb{R}\right)$ and the rotation angle is $2k\pi+\frac{\pi}{6}$(choice of right-handed rule), the homogeneous matrix of which can be conversely obtained via equation~\eqref{3D homogeneous roation} by normalizing the axis direction and set $t$ as any convenient specific real value, e.g., $0$. Do not forget to check the orthogonality of the upper-left $3\times 3$ submatrix up to a nonzero constant, which is a natural result by the isometric constraint, otherwise the verification may fail in~\eqref{eqn:3D rotation example b}.

The exactly equivalent to Rodrigues rotation form~\cite[pp583-585]{Hartley} when $a^2+b^2+c^2=1$:
\begin{gather}\label{eqn:3D homogeneous roation}\renewcommand\arraystretch{1}
\boldsymbol{R}^{3D}\left(x_0,y_0,z_0,a,b,c,\theta\right)=\boldsymbol{I}+\left(\sin\theta\cdot\mathscr{A}_2+ \left(1-\cos\theta\right)\cdot\mathscr{A}_2^2\right)\cdot \mathscr{T}_4
\end{gather}
\begin{flalign*}\renewcommand\arraystretch{1}
=\left[
\begin{array}{cccc}
 1 & 0 & 0 & 0 \\
 0 & 1 & 0 & 0 \\
 0 & 0 & 1 & 0 \\
 0 & 0 & 0 & 1 \\
\end{array}
\right]+
\left(\sin\theta{\left[
\begin{array}{cccc}
 0 & -c & b & 0 \\
 c & 0 & -a & 0 \\
 -b & a & 0 & 0 \\
 0 & 0 & 0 & 0 \\
\end{array}
\right]}+\left(1-\cos\theta\right){\left[
\begin{array}{cccc}
 0 & -c & b & 0 \\
 c & 0 & -a & 0 \\
 -b & a & 0 & 0 \\
 0 & 0 & 0 & 0 \\
\end{array}
\right]^2}\right)\cdot \left[
\begin{array}{cccc}
 1 & 0 & 0 & -{x_0} \\
 0 & 1 & 0 & -{y_0} \\
 0 & 0 & 1 & -{z_0} \\
 0 & 0 & 0 & 1 \\
\end{array}
\right]
\end{flalign*}


The previously discussed rotation formulation can be readily adapted for scenarios where nonhomogeneous coordinates are more advantageous. Consider a rotation axis defined by the unit vector $\boldsymbol{\hat{n}}=\left(a,b,c\right)^T$, with the condition $a^2+b^2+c^2=1$, and a point $\boldsymbol{p}=\left(x_0,y_0,z_0\right)^T$ located along this axis. In the context of 3D Euclidean vector algebra, the rotation of a vector $\boldsymbol{v}=\left(x,y,z\right)^T$ to $\boldsymbol{\hat{v}}=\left(x',y',z'\right)^T$ through an angle $\theta$ can be expressed as follows:


\begin{flalign}\renewcommand\arraystretch{1}\label{eqn:3D rotation matrix vector Euclidean}
\boldsymbol{\hat{v}}=\boldsymbol{v}+\left(\boldsymbol{I} \;\sin\theta +\left(1-\cos\theta\right) \left[
\begin{array}{ccc}
 0 & -c & b \\
 c & 0 & -a \\
 -b & a & 0 \\
\end{array}
\right]\right)\cdot \left[
\begin{array}{ccc}
 0 & -c & b \\
 c & 0 & -a \\
 -b & a & 0 \\
\end{array}
\right]\cdot \left(\boldsymbol{v}-\boldsymbol{p}\right)
\end{flalign}


Gimbal lock occurs when three consecutive rotations (not necessarily Euler rotations) are performed around axes that intersect at a common point and are mutually perpendicular, and the second rotation angle is $\pm\frac{\pi}2$. To avoid gimbal lock numerically, different methods can be employed, such as keeping the second rotation angle far enough from $\pm\frac{\pi}2$, introducing slight but enough offsets in the rotation axes to prevent them from intersecting, or allowing for small but enough deviations from perpendicularity among the axes. The matrix-vector form of general 3D rotation in equation~\eqref{eqn:3D rotation matrix vector Euclidean} allows for the convenient and efficient implementation of any of the aforementioned solutions to the gimbal lock problem.

\newpage

\def\arraystretch{1.5}

\begin{sidewaystable}
\centering
\caption{\large Classification and Definitions of Geometric Transformations Which are Stereohomology}
\label{classification table}
\resizebox{23cm}{!} {
 \begin{tabular}[t]{ccccccrr}%
\hlinewd{2pt}
\; No.\; & \tc{$S$ vs. $\pi$} & \tc{ \(\displaystyle\begin{casesBig} {\textrm{ Transformation}}{\textrm { matrix property}}\end{casesBig}\)} & \tc{Property of $\pi$} & \tc{ Property of $S$ }& {Transformation matrix formula\(\displaystyle(\lambda=1)\)} & \tc{Definition of transformation} \tabularnewline
\hline
1& $S\notin\pi$ & \tc{Singular} & \tc{Regular} & \tc{Regular} & & Central Projection {\color{red}\checkmark}\tabularnewline
2& $S\notin\pi$ & \tc{Singular} &\tc{Regular} & \tc{Infinite} &   \(\displaystyle\mathscr{T}\left ((s),(\pi); \lambda \right )=
 \lambda\cdot{\boldsymbol I} -\lambda\cdot \frac{{(s)\cdot(\pi) ^{\scriptscriptstyle \top}}}{{(s)^{\scriptscriptstyle \top}  \cdot (\pi)} } \)  & Oblique \& Orthographic Parallel Projection{\color{red} \checkmark} \tabularnewline
3& $S\notin\pi$ & \tc{Singular} & \tc{Infinite} & \tc{Regular} & & Direction {\color{red} \checkmark} \tabularnewline
\hline
4& $S\notin\pi$ & \tc{Nonsingular} & \tc{Regular} & \tc{Regular} & & {Space homology}{\color{red} \checkmark}\tabularnewline
5& $S\notin\pi$ & \tc{Nonsingular} & \tc{Regular} & \tc{Infinite} &  \(\displaystyle\mathscr{T}(\ (s),(\pi) ;\rho, \lambda)\:\mathop =\;{\lambda\cdot\boldsymbol I} + (\rho - \lambda) \cdot \frac{{(s)\cdot(\pi) ^{\scriptscriptstyle \top} }} {{(s)^{\scriptscriptstyle \top} \cdot (\pi) }} \) & Oblique \& Orthographic Elementary Scaling {\color{red} \checkmark} \tabularnewline
6& \tc{$S\notin\pi$} & \tc{Nonsingular} & \tc{Infinite} & \tc{Regular} & & Central Dilation {\color{red} \checkmark} \tabularnewline
\hline
7& $S\notin\pi$ & \tc{$\begin{casesBig} {\textrm{ Nonsingular}}{\textrm {\&Involutory}}\end{casesBig}$} & \tc{Regular} & \tc{Regular} & & {Involutory space homology} {\color{red} \checkmark}\tabularnewline
8& $S\notin\pi$ & \tc{$\begin{casesBig} {\textrm {Nonsingular}}{\textrm {\&Involutory}}\end{casesBig}$} & \tc{Regular} & \tc{Infinite} & \(\displaystyle\mathscr{T}\left ( (s),(\pi); \lambda \right )=
 \lambda\cdot{\boldsymbol I} - 2\lambda \cdot {\huge \frac{(s)\cdot(\pi)\! ^{\scriptscriptstyle\top}} {(s)\!^{\scriptscriptstyle\top} \! \cdot (\pi) }} \) & Skew(Oblique) \& Orthographic Reflection {\color{red} \checkmark}\tabularnewline
9& $S\notin\pi$ & \tc{$\begin{casesBig} {\textrm {Nonsingular}}{\textrm {\&Involutory}}\end{casesBig}$} & \tc{Infinite} & \tc{Regular} & & Central Symmetry {\color{red} \checkmark}\tabularnewline
\hline
10& $S\in\pi$ & \tc{Nonsingular} & \tc{Regular} & \tc{Regular} & & {Space elation}{\color{red} \checkmark}\tabularnewline
11& $S\in\pi$ & \tc{Nonsingular} & \tc{Regular} & \tc{Infinite} &  \(\displaystyle \mathscr{T}((s),(\pi) ;\lambda,\mu )=
{\lambda\cdot\boldsymbol I} + \frac{ \mu \cdot
(s)\cdot{(\pi)}^{\scriptscriptstyle \top} } {\sqrt
{(s)^{\scriptscriptstyle \top}\!\!
\cdot\! (s)\!\cdot\!(\pi) ^{\scriptscriptstyle \top}\!\!\cdot\! (\pi) } }\) & Shearing {\color{red} \checkmark} \tabularnewline
12& $S\in\pi$ & \tc{Nonsingular} & \tc{Infinite} & \tc{Infinite} & & Translation {\color{red} \checkmark} \tabularnewline
\hlinewd{1pt}
\end{tabular}
}
\end{sidewaystable}

\end{document}